# Video Generative Adversarial Networks: A Review

NUHA ALDAUSARI

University of New South Wales, Australia

Princess Nourah bint Abdulrahman University, KSA

ARCOT SOWMYA

University of New South Wales, Australia

NADINE MARCUS

University of New South Wales, Australia

GELAREH MOHAMMADI

University of New South Wales, Australia

**ABSTRACT**

With the increasing interest in the content creation field in multiple sectors such as media, education, and entertainment, there is an increasing trend in the papers that uses AI algorithms to generate content such as images, videos, audio, and text. Generative Adversarial Networks (GANs) in one of the promising models that synthesizes data samples that are similar to real data samples. While the variations of GANs models in general have been covered to some extent in several survey papers, to the best of our knowledge, this is among the first survey papers that reviews the state-of-the-art video GANs models. This paper first categorized GANs review papers into general GANs review papers, image GANs review papers and special field GANs review papers such as anomaly detection, medical imaging, or cybersecurity. The paper then summarizes the main improvements in GANs frameworks that are not initially developed for the video domain but have been adopted in multiple video GANs variations. Then, a comprehensive review of video GANs models is provided under two main divisions according to the presence or non-presence of a condition. The conditional models then further grouped according to the type of condition into audio, text, video, and image. The paper is concluded by highlighting the main challenges and limitations of the current video GANs models. A comprehensive list of datasets, applied loss functions, and evaluation metrics is provided in the supplementary material.

## 1 INTRODUCTION

The field of Computer Vision mainly deals with two types of data, namely images and videos, and these data can be used in many real-life applications for data generation, editing and classification. Data generation has gained significant attention since Ian Goodfellow released a model called Generative Adversarial Networks (GANs) in 2014 [1]. According to Google Scholar, there is an upward trend since the mid 2010's in publications when specifying "generative adversarial networks" as a search keyword, as demonstrated in Figure 1.

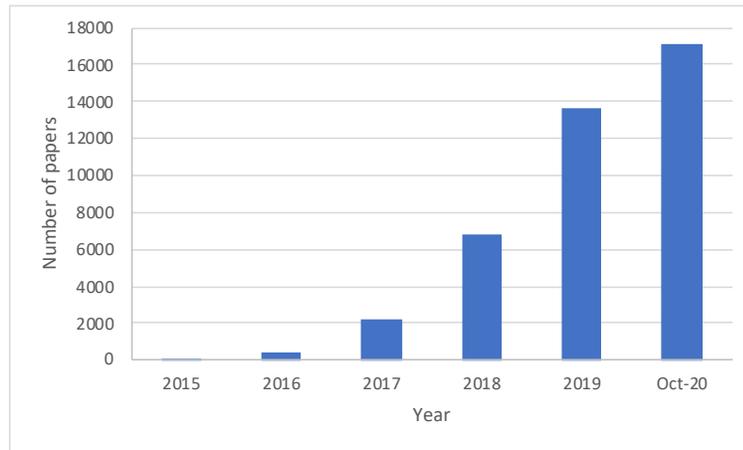

Figure 1: line chart represents the number of papers published in each year according to Google Scholar from 2015 to Jun 2020 (vertical axes)

The goal of generative models, in general, is to generate new data points that conform to the distribution of the training dataset. To accomplish this goal, GANs consists of two networks, one named the generator that gets a random noise vector as input and outputs images. The second network is the discriminator that differentiates between the real training images and fake ones created by the generator. In other words, the discriminator D classifies real images x as D(x)=1 and the fakes ones as D(x)=0. The networks are trained in an adversarial manner to reach the Nash equilibrium, which is an optimal state where D(x) = ½ for each image x, which means that the discriminator is not able to differentiate between real and fake samples [1].

One reason behind the success of GANs frameworks is that they overcame some of the limitations of other generative models such as Variational Autoencoders (VAEs) [2]. For example, GANs frameworks produce sharper images compared to VAEs. The reconstruction loss function of VAEs is a pixel-wise similarity metric, while that of GANs is a semantic loss function [3, 4]. The issue with element-wise measures is that they do not align with the human visual system in the image domain; in other words, there are images that reflect a high element-wise error, however humans cannot distinguish between these images, and vice versa. The adversarial training in GANs facilitates building the reconstruction loss of the generator implicitly through the back-propagating gradients of the discriminator model that is responsible for differentiating between real images and fake ones.

GANs have been applied successfully on images and produce 1024*1024 images [5] that humans cannot differentiate from photographed images. Because a video is a sequence of images, it is also possible to employ GANs in the video domain. However, the main challenge in synthesizing videos is that a video contains multiple images rather than a single image. Besides, a video constitutes multimodal data with different aspects such as speed, motion, picture and soundtrack. Moreover, the temporal dimension and the dependency between the frames adds to the challenges of generating videos. Although dealing with video in GANs can be more complicated than handling images, many research works have already applied GANs to video datasets, with the first attempt by Vondrick et al in 2016 [6]. This was followed by others to leverage the use and value of applying GANs to a video dataset. Since then, several review papers have included a discussion of video GANs models [7-10]. As far as is known, this paper is the first attempt to review video GANs models more extensively.

This paper is structured as follows: an overview of the main enhancements of GANs models and their variations are presented in section 2. Other GANs review papers are summarized in section 3, and gaps in the literature identified. Video GANs models are categorized into unconditional and conditional models in section 4 and each category reviewed. Finally, section 5 concludes the paper.

## 2 RECENT ADVANCES IN GANS

This section presents remarkable GANs variations that are employed in video GANs models that are detailed in section 4. The original GANs model [1] is presented in section 2.1. Three GANs frameworks with conditional settings are described in section 2.2. In section 2.3, an overview of another type of GANs named convolutional GANs [11] is provided. In sections 2.4 and 2.5, other types of GANs called BigGAN and Wasserstein GAN respectively are described. A visual comparison of the overall architectures is illustrated in Figure 2.

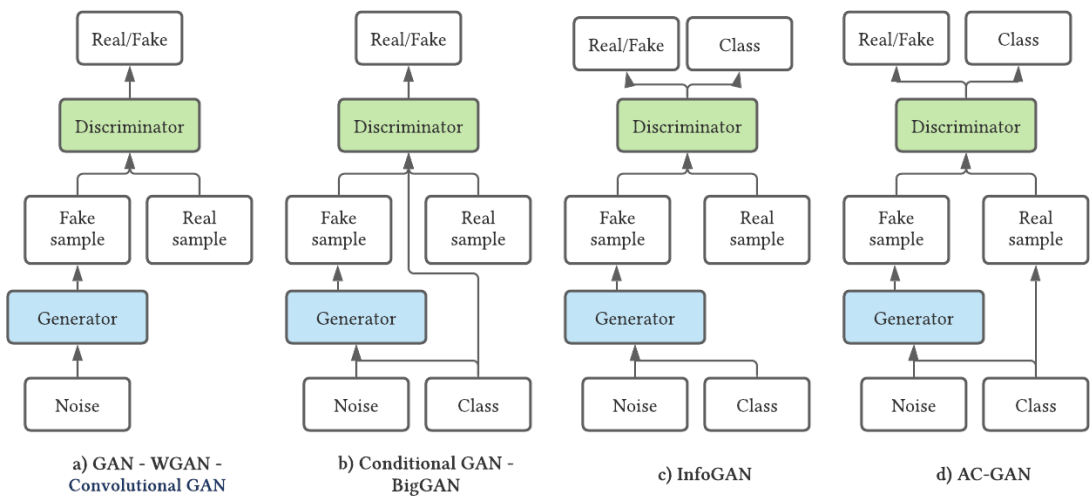

Figure 2: Architecture of GANs variants. From left to right: a) is that of GANs, WGAN, and convolutional GANs, which share the same architecture. b) is conditional GAN, and BigGAN is based on it. c) and d) are of InfoGAN and AC-GAN respectively.

### 2.1 Vanilla GANs

GANs framework was first introduced by Ian Goodfellow and his colleagues in 2014 [1]. Since then, GANs has attracted a lot of attention and played a significant role in the field of generative models. This is because GANs surpass other generative models with its unprecedented ability to generate new samples. GANs have multiple applications in different domains, including text, images and sounds. Yann LeCun referred to GANs as "the coolest idea in machine learning in the last twenty years" [12].

The abbreviation GANs is based on three words: "Generative" means synthesizing new data based on training sets; "Adversarial" indicates that the two components of GANs, namely the generator and the discriminator, contest against each other, while the word "Networks" illustrates that the model consists of (two) networks. The networks could be fully connected neural networks, convolutional neural networks, recurrent neural networks, long short term memory neural networks, autoencoders or any combination thereof.

GANs consist of two networks competing in a minimax game, as illustrated in Figure 2a. One network, called the generator, takes a random noise vector as input, and produces new instances by learning to follow real data distribution. On the other hand, the discriminator accepts training data and the data that is synthesized by the generator. The discriminator is a classification network that is supposed to classify real training data samples as "1"s and the generated data points as "0"s. The objective of the generator is to generate samples that cannot be distinguished from the real data samples by the discriminator. On the contrary, the discriminator targets the discrimination of real data samples from fake ones via classification. The two networks compete in a minimax game to improve each other's performance. The ultimate goal for both networks is to reach the Nash equilibrium, which is a state where neither of the networks can improve by changing their parameters. In practice, however, it is difficult to find the Nash equilibrium [12].

The objective function for the generator (G) and the discriminator (D), as previously stated, is defined as a loss function (L) [1]:

$$min_G \ max_D \ L(D,G) \ = \ E_{x \sim p_r(x)}[log \ D(x)] \ + \ E_{z \sim p_z(z)}[log(1 \ - \ D(G(z)))] \quad (1)$$

In equation (1), $E_p$ is the expectation with respect to a distribution p, $p_r$ refers to the distribution of the real data, and $p_z$ is the distribution of the input noise vector $z$. The discriminator is trained to recognize the real samples $x$, and produce high values close to one. Therefore $E_{x \sim p_r(x)}[log \ D(x)]$ should be maximized. Meanwhile, the discriminator is also trained to recognize the fake samples $G(z)$ and produces low values close to zero, which means maximizing $E_{z \sim p_z(z)}[log(1 \ - \ D(G(z)))]$. In contrast, the generator needs to generate samples $G(z)$ that are similar to the real samples in order to fool the discriminator $E_{z \sim p_z(z)}[log(1 \ - \ D(G(z)))$.

### 2.2 Conditional GAN

In vanilla GANs [1], the model is unable to control the type of the generated samples. The generated data points could be from any category of the training data distribution. When sampling occurs, the generated samples might not represent all possible variations of the training data. In contrast, conditional GAN (CGAN) [13] adds a condition to both the generator and the discriminator, as shown in Figure 2b. The condition might be a class, text or any other type of data, and the generated data is expected to match the condition. The loss function for CGAN is a modified version of the GANs loss function in equation (1):

$$min_G \ max_D \ L(D,G) \ = \ E_{x \sim p_r(x)}[log \ D(x|c)] \ + \ E_{z \sim p_z(z)}[log(1 \ - \ D(G(z|c)))] \ [13] \quad (2)$$

where c is the condition added to the model.

Information Maximizing GAN, InfoGAN [14] for short, uses a slightly different approach to control the generation process. As illustrated in Figure 2c, the input in InfoGAN [14] to the generator is a noise vector along with another variable. Unlike the condition in CGAN, the variable vector in InfoGAN is unknown. The purpose of using the variable is to control specific properties in the generated samples by maximizing the mutual information between this variable and the generated samples. Through training InfoGAN, the generator learns how to disentangle certain properties in the generated samples in an unsupervised manner through another model called the auxiliary model. The auxiliary model shares the same parameters as the discriminator. The aim of the auxiliary model is to predict the properties that are disentangled while the discriminator's purpose is to distinguish real samples from fake ones. After training, the generated samples can be controlled by specifying some features that are learned such as colour, shape, rotation in image samples or class [14]. The loss function for InfoGAN is defined as follows:

$$min_G \ max_D \ L(D,G) \ - \ \lambda I \ (c; \ G \ (z,c)) \ [14] \quad (3)$$

The loss function is the same as CGAN, but with an additional term $\lambda I \ (c; \ G \ (z,c))$ to represent the mutual information loss between the variable c and the generated samples [14].

Auxiliary Classifier GAN (AC-GAN) [15] is another framework under conditional GANs-based architectures. AC-GAN frameworks share characteristic with InfoGAN and another with CGAN. In terms of the similarity with CGAN, a condition is fed into the generator, which can be a text, class or any other type of data. However, the condition is not an input for the discriminator. The common factor between AC-GANs and InfoGAN is that there is an auxiliary classifier that outputs the class of the input sample. The differences in the overall architecture between CGAN, InfoGAN, and AC-GAN are demonstrated in Figure 2b, 2c, and 2d respectively. The loss function of AC-GAN is divided into two terms: one for evaluating the predicted class and another one to discriminate fake from real samples.

The reviewed papers under conditional video generation (see section 4.2) use one of the conditional GANs-based architectures. The reason behind choosing the conditional architecture is that conditions can enhance the network stability and provide higher quality samples [13].

## 2.3 Convolutional GANs

Vanilla GANs [1], considered as the simplest type of GANs, uses the Multi-Layer Perceptron (MLP) in both the generator and the discriminator. However, one of the main disadvantages of vanilla GANs is that the training process is not stable [12]. One of the possible solutions to this problem is to use a convolutional neural network (CNN) instead. In convolutional GANs [11], the generator uses a deconvolution structure, while the discriminator applies convolutional layers to classify generated images from the real images. While the type of networks in the generator and discriminator are different in vanilla GANs and convolutional GANs, the overall architecture in vanilla GANs and convolutional GANs are the identical (see Figure 2a).

A significant number of recent GANs frameworks adopt CNN in their generators, discriminators, or in both. This is because the CNN in Convolutional GANs outperforms MLP in vanilla GANs in terms of the performance, quality of the resulted samples (usually images) and stability of the training process.

## 2.4 BigGAN

The GANs frameworks discussed so far when applied in image domain produce low resolution images as output, i.e 64×64 or 128×128 images. In contrast, BigGAN [16] scales up the resolution by increasing the number of parameters and scaling up the batch size. Therefore, training BigGAN on ImageNet results in 256×256 and 512×512 images. The architecture of BigGAN is based on the self-attention GAN (SAGAN) [17]. The main advantage of SAGAN is that it focuses on different parts of the image by introducing an attention map that is applied to the feature maps in the deep convolutional model architecture.

## 2.5 Wasserstein GAN

Another variation of GANs that has been used in video generation is called Wasserstein GAN (WGAN). In vanilla GANs [1], the discriminator attempts to classify real data points from fake ones. However, WGAN's discriminator [18] is a "critic", whose responsibility is to assign a score that represents the distance between the distribution of the observed real data and the distribution of observed fake samples. WGAN uses Wasserstein distance instead of the Jensen-Shannon (JS) divergence and Kullback-Leibler (KL) divergence that are used in other generative models, and there is substantial improvement in the quality of the generated images and in the training stability.

## 3 RELATED WORK

As the study of GANs is accelerating rapidly, there are constantly new GANs frameworks that were not covered in the existing review papers. The timeline of survey papers found by searching Google Scholar for the keywords "overview of generative adversarial networks", "survey of generative adversarial networks", and "review of generative adversarial networks", as well as the papers cited in the retrieved papers, is illustrated in Figure 3. The timeline also includes the publication dates of the review papers, which will be discussed in this section, besides the publication dates of notable GANs frameworks discussed in section 2. While some of the review papers provide overviews of the state-of-the-art GANs, others focused on GANs for a specific domain (e.g. image generation). Reviews of GANs for general visual image datasets exceed other specialized GANs reviews, including those in cybersecurity, anomaly detection and medical imaging (see below). As shown in Figure 3, the red diamonds that represent general reviews of GANs dominate other categories.

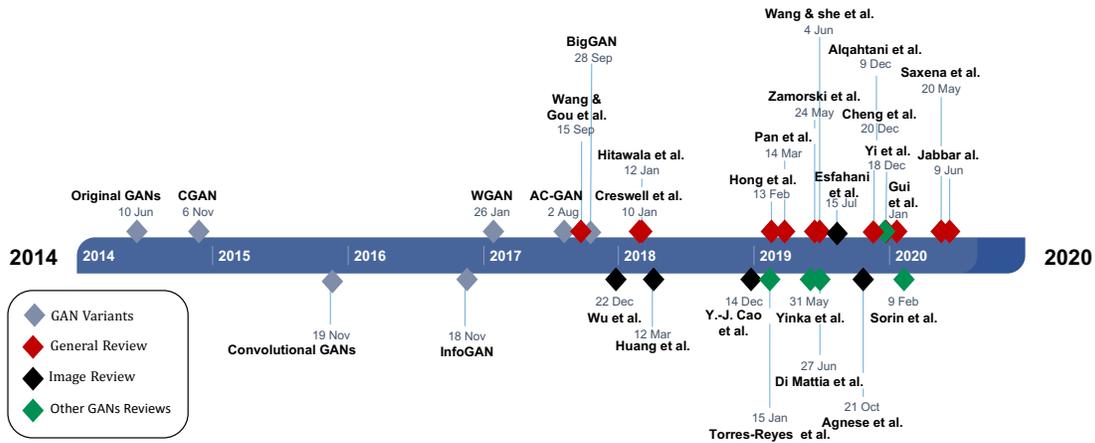

Figure 3: Timeline of the review papers in section 3, along with recent GANs advances in section 2 (light blue)

## 3.1 General GANs

The majority of the survey papers discuss GANs in general [7-10, 19-23]. One of the first attempts to review general GANs was in late 2017 [22]. Another general GANs survey was published in early 2018 that draws an analogy between GANs concepts and signal processing concepts to facilitate the understanding of GANs from a signal processing perspective [20]. Around the same time, Hitawala et al. [23] provided a review of the main improvements in the GANs frameworks. In 2019 and 2020, there was an increase in the number of papers that review GANs in general [7-10, 19, 21, 24-26]. It is worth mentioning that a more comprehensive and lengthy general review is available [19], and Cheng et al. [25] conducted comparative qualitative experiments on the mainstream GANs applied to the MNIST dataset [27], in which AC-GANs obtained the top classification accuracy. Saxena et al. [26] concentrate on models that address GANs disadvantages such as training instability and model collapse by modifying the architecture, the loss function or the optimization method.

### 3.2 Image GANs

Since the original GANs frameworks were initially built upon images, there is no doubt that the number of GANs applications in the image domain surpasses other areas such as text, voice and video. Multiple reviews [28-32] focus on image synthesis, even though the major proportion of the general GANs reviews mentioned above, are also in the image domain. Huang et al. [29] categorize the image synthesis GANs frameworks into three types based on the overall architecture. These include direct architectures based on vanilla GANs, hierarchal models and iterative models consisting of multiple generators and discriminators. While each generator in an hierarchal architecture is tasked to deal with a different aspect of the disentangled representations of training images, the goal of a generator in the iterative models is to refine the quality of the generated images. Wu et al. [30] place image GANs models in two main categories, namely conditional image synthesis models and unconditional image synthesis models. In the unconditional models, different network modules handle texture, image super resolution, image inpainting, face synthesis and human synthesis. A comparative study [31] of image GANs frameworks was conducted using two datasets: MNIST[27], and Fashion-MNIST[33]. The paper reviews different applications of GANs such as style transfer, image inpainting, super-resolution and text to image. Similar applications are also reviewed elsewhere [28], with additional applications such as face ageing and 3D image synthesis. Moreover, Agnese et al. [32] direct attention to GANs models that are conditioned on text and produce images. Some of these models fall under semantic enhancement GANs, where the main goal is to ensure that text is semantically coherent with the generated image. Another category focusses on producing high resolution images conditioned on text. An additional category is for ensuring the diversity of synthesized images based on input text. The last type is text to video models that consider the temporal dimension of the training samples.

### 3.3 Other GANs

GANs have been applied successfully in other areas including the medical field, with medical GANs frameworks reviewed [34, 35]. Yi et al. [34] provide illustrations of juxtaposed GANs variation architectures. In addition, medical imaging GANs models are categorized based on the aim, such as quality improvement of the synthesized images, data augmentation, segmentation, classification, registration and object detection. Another review [35] specializes in radiology imaging based on 33 papers. GANs models facilitate synthesis of new radiology images, improving the quality of existing ones, converting radiology images from one type to another and localizing a specific object.

Anomaly detection, which deals with finding data samples that deviate from the normal, has also taken advantage of GANs models. A survey paper [36] reviews existing GANs models that contribute to identifying anomalies, and re-implements the reviewed models to further verify the effectiveness of such models.

The field of synthesizing and enhancing audio using GANs architectures has also been reviewed [37]. In the audio generation models, audio can be generated form a noise vector or text, while in audio enhancement GANs models, the model input is a noisy audio signal while the output is a refined signal. Audio GANs models are usually trained on audio spectrogram representations or raw audio waveforms.

Yinka-Banjo et al. [38] review studies that utilize GANs frameworks in cybersecurity systems. Such models are used for protection or attack purposes. In the case of protection, GANs models synthesize new poisoning samples, and the cybersecurity system learns how to protect itself against such samples that might cause an attack. GANs systems can also be used to spoof other GANs cybersecurity systems by creating adversarial samples that are similar to the real authorized training data points.

Besides general GANs reviews, there are also domain-specific reviews that focus on image, audio and medical imaging as listed above. In addition, several review papers are focussed on specific tasks such as anomaly detection and cybersecurity. However, no systematic review of video GANs has been performed, which have very specific characteristics due to their temporal dimension and the necessity to maintain temporal cohesiveness. The next section contributes to filling this gap and pays close attention to GANs models that generate videos and capture their temporal behavior.

## 4  VIDEO GANS

Video GANs models vary depending on the condition settings. While at one end, there are video GANs frameworks that are not supplied with conditional signals, discussed in section 4.1, other models are conditioned on audios, texts, semantic maps, images and videos, as discussed in sections 4.2.1, 4.2.2, 4.2.3, and 4.2.4 respectively. In all these categories, the overall architecture can be divided roughly into three subcategories. First are the GANs frameworks that use RNN architectures to address the time-series nature of video data [39, 40]. The second are progressive video GANs models [41, 42] in which initial frames are first generated, and the generated data is fed into another generator to produce an enhanced result. The third are the two-stream architecture video GANs [6, 43], where each stream considers a different aspect of the video. The following sections review the video GANs models based on the presence or absence of input condition and the variations within each category.

### 4.1  Unconditional video generation

This section is a review of unsupervised GANs frameworks in the video domain, and a summary of these frameworks, the datasets and evaluation metrics used can be found in Table S4. So far, the output videos produced by these frameworks are short and have low-quality frames due to the lack of any information provided as a condition with the videos during the training phase. Although such models produce low-quality videos, the unconditional models have become the foundation for conditional frameworks. For instance, MoCoGAN [40] architecture, which is an unconditional model, is used in Text-Filter conditioning Generative Adversarial Network (TFGAN)[44] and storyGAN [39], both of which are conditional models.

Video-GAN (VGAN) [6] was the first attempt to generate videos using GANs. The generator consists of two convolutional networks: the first is a 3D spatio-temporal convolutional network that captures moving objects in the foreground, while the second is a 2D spatial convolutional model for the static background. The generated frames from the two-stream generator are combined, and then fed to the discriminator to distinguish real videos from the fake ones.

In VGAN [6], the foreground stream captures the foreground objects and their motions. However, the foreground layer in the generated result usually contains some flaws in temporal or spatial aspects. Flow and Texture Generative Adversarial Networks (FTGAN) [45] adds optical flow for representing the object motion more effectively. FTGAN follows a progressive architecture that starts with a GANs framework to captures the optical flow, followed by another GANs model to generate the texture that is conditioned on the result of the previous optical flow GANs, and produces the desired frames. Both texture and flow generators in FTGAN adopt VGAN structure by separating the foreground from the background, and setting the background to zero for the flow generator.

VGAN is based on disentangling the foreground from the background. Similarly, Motion Content GAN (MoCoGAN) [40], which is another type of unconditional video generator, separates the content from the movements to provide more control over these components. While VGAN [6] and FTGAN [45] map a video to a point in the latent vector, the MoCoGAN framework traverses N latent points, one per frame, where each vector can be decomposed into the motion vector and content vector. Therefore, it consists of an N-to-N RNN that accepts N random variables and produces N latent motion vectors. The motion vectors are combined with a fixed content vector for all N motion variables and fed to the generators to synthesize N images, each of which is a frame in the generated video. The generated images and videos are evaluated using two discriminators: one for the images, and the other for the generated video.

Similar to MoCoGAN [40], Temporal Generative Adversarial Nets (TGAN) [46] use N latent vectors for N frames. However, the main difference between MoCoGAN and TGAN is that a frame is generated for every latent vector in TGAN, which is different from the disentangled content and dynamic latent vectors in MoCoGAN [40]. Another difference is that MoCoGAN utilizes RNN structure for the generators while in TGAN, there are N 2D image deconvolutional generators to produce N frames. The resultant frames along with the videos in the training set are fed into the 3D convolutional discriminator. TGAN employs WGAN and fulfills K-Lipschitz constraint by proposing a parameter clipping method called singular value clipping using WGAN to provide stable training. Temporal GAN v2 (TGANv2) [47] focusses on a training technique based on understanding the relationship between the resolution of the generated images and the computational cost. The main cause of the increase in computational cost is the end part of the generator. This is because the spatial resolution / feature map is increasing when moving forward in the generator net. Since TGANv2 is an unsupervised model, it is essential to supply the model with a larger batch size in order to generalize properly. A subsampling layer is introduced to reduce the batch size by stochastically sampling videos within a mini-batch, and then sampling a frame within each chosen video. Applying the subsampling technique multiple times in the generator network facilitates reduction in the size of the mini-batch.

Dual video discriminator GAN (DVD-GAN)[48] expands BigGAN [16] capabilities in the video domain to produce 48 high quality images up to 256*256 based on complex datasets such as Kinetics human action dataset. DVD-GAN is trained on the entire dataset, Kinetics, and this is not the case in prior works [42, 44] that use only a subset and pre-processed samples. Similar to MoCoGAN [40], there are two discriminators to deal with the temporal and spatial aspects of a video.

## 4.2 Conditional video generation

There are several works that employ a conditional signal in GANs to direct the process and control modes of the generated data;The condition may be audio signal, text, semantic map, image or video. The following subsections review conditional GANs based on the condition type.

*4.2.1 Speech to video synthesis*

This subsection discusses the GANs frameworks that are used to synchronize speech audio with facial movements, and Table S5 summarizes speech-to-video synthesis models.

Lips movement generation frameworks were the initial attempt at synchronizing a moving head with audio. Chen et al. [49] proposed a model that encodes the starting image and audio file. Then, the encoded features are combined and used as input to a decoder to generate videos. The synthesis videos are evaluated using a three-stream discriminator.

While Chen et al. [49] consider only lips, other works [50-53] study the synchronization between audio and the entire face. Jalalifar et al. [53] proposed a progressive framework that combined LSTM with CGAN. The purpose of LSTM is to extract the landmarks of the mouth region. Given the landmarks as a conditional setting, CGAN synthesizes a synchronized talking face for the audio signal. Vougioukas et al. [50] converts an audio waveform file to a synchronized spoken person, without an intermediate step for extracting the landmarks as in Jalalifar et al. [53]. In this model, given the initial frame, a temporal GANs model with two discriminators, one for the frame level and the other for the sequence level, are trained to produce synchronized videos for the audios.

The approach of disentangled representations has been considered [51, 52]. The intuition behind the Disentangled Audio-Visual System (DAVS) [51] is to decouple the talking head information into person-related information and speech-related information in order to overcome the blurry and incoherent videos generated from a plain speech file. DAVS is able to generate videos of talking faces based on audio files or other video files, while other studies [50] [52] are conditioned only on the audio file. DAVS is able to retarget face movements from one video to another.

Mittal et al. [52] decoupled the audio into three aspects: content, emotion and noise, contrary to other works [51],[50],[49] that utilize the audio file without any per-possessing step. Decomposing audio representations using Variational AutoEncoder (VAE) at the first stage facilitates discarding of the background noise that appears in real world recorded datasets. Moreover, eliminating emotion from content reduces the effect of emotion on the generated videos. In the second stage, the content component of the disentangled audio with an image from the video is fed into the generator to produce a frame. There are two discriminators in this framework: a frame level discriminator and a video level discriminator.

*4.2.2 Text to video synthesis*

This subsection considers GANs-based frameworks that aim to produce videos according to a conditional text. These frameworks have two main purposes. The first is to maintain semantic consistency between the condition and the generated video. The second purpose is to generate realistic quality videos that preserve the coherence and consistency within the frames. The main text to video GANs frameworks are summarized in Table S6.

Temporal GANs conditioning on captions (TGANs-C) framework [54] first encodes the text using an LSTM based encoder. The output of the sentence encoder is concatenated with a noise vector and then given to the generator, which is a 3D deconvolution network. The model has three discriminators for the video level, frame level and the motion level, to ensure that adjacent frames have coherent motion.

While videos in TGANs-C [54] are generated using a single generator, the GANs framework proposed elsewhere [42] generates videos progressively using multiple generators in several stages. Firstly, the conditional variational autoencoder that is conditioned on encoded text produces the initial image. This initial image provides an overall representation, which may be the background image, the colours of the image and its structure. The initial image with the encoded text is an input to a CGAN to generate higher quality images.

The original method used in CGAN [13] to incorporate a condition is to concatenate the condition and the noise vector; this method is also applied in text to video GANs frameworks [42, 54]. However, TFGAN [44] introduces a multi-scale text-conditioning method, where the text features are extracted from the encoded text to generate convolution filters. Then, the convolution filters are input to discriminator network to facilitate strengthening of the associations between the texts and the videos.

In video generation models based on text reviewed so far [42, 44, 54], a model synthesizes a video according to one conditional sentence per video. In contrast, storyGAN [39] is a story visualization model that is conditioned on multiple sentences. StoryGAN [39] contains a context encoder and a story encoder. The story encoder encodes the entire story as a low dimensional vector that serves as an initial input to the context encoder. At each time step in the RNN context encoder, one sentence with concatenated noise is introduced along with encoded story vector to produce a Gist vector, which is combined information about a specific sentence and the story. The generator then accepts a Gist vector and produces an image. There is a discriminator for the image and another discriminator for the story. Different to other video generation frameworks, storyGAN pays less attention to the continuity of motion and instead focusses on the global consistency of the story.

*4.2.3 Semantic map to video synthesis*

This section represents video GANs frameworks that are conditioned on semantic maps. This section could potentially fall under section 4.2.5, video to video synthesis, since the frameworks are conditioned on videos. However, the conditioning videos are pre-processed into semantic maps first, and Table S7 provides a summary of the cited semantic map to video frameworks.

Video to video (vid2vid) [55] is a conditional GANs framework that converts semantic videos into frames. Semantic videos consist of semantic maps, where each map is a collection of segmented objects that are labeled with different colours. The semantic videos could be in the form of segmentation masks or boundaries. Few-shot-vid2vid [56] is an extension to vid2vid. Both vid2vid and few-shot-vid2vid share an overall architecture for the generator network that is conditioned on the previous frame, a previous semantic image, and the source semantic videos. The generator consists of three modules, namely W to extract the optical flow, M to predict the occlusion map, and H to generate the intermediate frames. The main difference between vid2vid and few-shot-vid2vid is that the module H in vid2vid has fixed weights, whereas the weights are dynamic in few-shot-vid2vid. An adaptive network with dynamic weights [56] facilitates generation of videos of unseen objects in the training dataset by providing multiple images of the object at test time. There are two discriminators in both architectures: a video discriminator and an image discriminator. The modules are trained in a progressive manner, which means that the number of frames and the quality of generated images increase gradually.

While the conditional signal in vid2vid and few-shot-vide2vid [55, 56] is a sequence of semantic maps, Pan et al. [57] only use one semantic label map. They claim that providing a single semantic map helps loosen the restrictions during the synthesizing process. To generate a video conditioned on a single semantic map, there are two phases. The first is an image-to-image phase that is conditioned on the semantic map to generate the initial frame with fine details. The second phase produces a video given the starting frame using conditional VAE.

*4.2.4 Image to video synthesis*

The main purpose of image-to-video GANs frameworks is to predict future frames based on a given frame, and Table S8 lists the reviewed image to video frameworks and some of their properties.

Early versions of image to video architectures [58-61] do not disentangle the representations of the training videos, resulting in blurriness in the synthesized videos. Mathieu et al. [58] made the first attempt to employ adversarial training in the video prediction domain. The generator is a multiscale network that is focussed on synthesizing coherent frames conditioned on front frames. The adversarial network solves the issue related to blurry frames that results from standard mean squared error loss function. Lee et al. [59] incorporate VAEs with GANs for a video prediction system. Combining VAEs with GANs was first performed in the image domain [4, 62]. The reason for using both GANs and VAEs networks is that a GANs model helps produce more realistic images, while VAE facilitates diversity in the generated images. Multi-Discriminator GAN (MD-GAN) [60] employs two consecutive GANs. While the first generates the content of the frames, the other GANs model refines the output of the first stage. Similar to MD-GAN [60], the model by Cai et al. [61] adopts a two-stage framework based on pose estimation. The first network generates human pose sequences of the conditioned type of pose, whereas the second stage network maps from the pose space to pixel space, given a reference image.

Several studies [43, 63-67] on frame prediction have shown that decomposing the information in videos enhances the quality of the synthesized videos. Motion Content Network (MCnet) [63] encodes motion and content using separate encoders in an unsupervised manner. The outputs of the previous stages are combined and decoded to produce the next frame. Walker et al. [64] also tackle content and motion in a progressive manner by utilizing two architectures, namely VAEs and GANs, as do Lee et al. [59]. The VAEs framework predicts future human poses and motions, while the GANs framework is conditioned on the motions to predict future frames as a pixel-level representation. Similar to Villegas et al.[63] and Walker et al. [64], both Sun et al.[43] and Liang et al.[65] attempt to separate the dynamics from the content by employing dual GANs. One of the GANs is dedicated to generating the content of a frame, while the other GANs model predicts the dynamics. The difference between the two architectures is that Liang et al. [65] use one encoder to encode the front frames, while TwoStreamVAN [43] encodes an initial frame with the content encoder and the motion map is encoded using the motion encoder. Using same approach as TwoStreamVAN, DRNET [66] has two encoder networks: one for the dynamic content and the other for time-independent content. The encoded representations are concatenated and fed into the decoder to predict the next frames. When decomposing motion from content to forecast a future frame, the motion component is detected from the input frames [43, 63-66]. However, Hu et al. [67] utilize a motion stroke, which is a continuous line that represents object motion, instead of extracting motion from the input frames. The model first encodes the initial image and the motion strokes. This is followed by an additional encoder to encode the encoded initial image, encoded motion strokes and features of the previous frame. Next, the generator is used to generate the sequence frames with two discriminators, one each for image level and sequence level.

*4.2.5 Video to video synthesis*

One of the major applications of video to video synthesis is object animation, where motion is retargetted from one object to another. Some GANs frameworks in this domain are limited to a specific dataset, while others can be applied in different domains, and Table S9 lists video animation frameworks.

Many works [68-70] share the main objective of converting a person's dance movements to those of another immature dancer. Zhou et al. [69] build their architecture in a progressive manner, where the first phase synthesizes frames for the immature dancer based on the pose of the mature dancer and body parts of the immature dancer. The second network provides more realism to the final output by fusing the target performer with a background and adding necessary shadows to combine the foreground with the background. Chan et al. [68] start with a pose detector network, whose result is inserted as an input to a GANs framework. Then, the GANs face framework is used to enhance realism. Yang et al.'s architecture [70] is similar to Kim et al.'s [71], where the model starts with disentangling the video representations of the source and target videos into distinct parameters that can be combined to produce retargeted video. Often ,the datasets [68] [70] are limited, with a small number of participants performing a wide range of movements. Thus, Chan et al's [68] computational module translates the poses more easily than Zhou et al.'s [69] , as the latter's dataset is collected from YouTube and it is not feasible to collect different poses of the same person.

Studies by Siarohin et al. [72] and Liu et al. [73] aim to generate a video that has similar action as an input video and containing an object similar to ones in an input image; in these models, the action is not restricted to a dance movement. When comparing the methods for pose extraction, it is important to mention that models by Zhou et al. Chan et al. [68, 69] are based on pre-trained networks for extracting subject movements. In contrast, MOviNg KEYpoints (Monkey-net) [72] is based on a self-supervised framework while Liu et al. [73] used 3D reconstruction software to rebuild the desired poses. In addition, other generation processes [56, 69, 72, 73] are conditioned on a specific human target, whereas Chan et al. [68] use a random human subject from the dataset in generated videos. Monkey-net is composed of three stages: the first stage extracts key points of two random images of the input video; the second stage is a motion prediction network that computes the optical flow of the result of the previous stage; and the last stage performs image synthesis using a Variational AutoEncoder. Liu et al. [73] start with extracting a 3D object from the static images and motion of the given video, and end with a CGAN framework to produce realistic frames.

vid2Game [74] follows a different approach to control the human subject in generated frames. The synthesized images are conditioned on a low dimensional signal such as joystick movements. There are two networks: the first is Pose2pose, which generates a new pose given the current one and the control signal using an autoencoder. The control signal is input to the residual blocks in the middle of the autoencoder to facilitate smooth motion. The second network is the Pose2frame network, which generates a mask and an image. The image is combined with the mask to produce a frame. In other work [55, 56, 68, 69], the generator network is conditioned on the pose extracted from real images of the training dataset, while in vid2Game [74], the network is conditioned on synthesized poses. Clearly, working with poses extracted from synthesized frames requires more effort compared to working with poses from the training images, as artefacts in the generated poses need to be dealt with.

While some studies [64, 65, 68, 95] retarget the motion of an individual body to another individual, others focus only on the head [71, 75]. Deep video portraits [71] outputs transferred motion video that can be modified by changing adjustable parameters such as head pose and facial expression. Deep video portraits translate both source and target videos to low dimensional parameter vectors that are adjusted to input to a rendering video. ReenactGAN[75] extracts the face boundary as an initial step, then the source boundary is aligned with the target boundary to produce target video animation. Wu et al. [75] claim that ReenactGAN is better at representing minor changes in the face between frames than other frameworks [71].

Recycle-GAN [76] is another video retargetting application that takes a different approach. While other frameworks transfer motion from one object to another[64, 65, 68, 95][71, 75] , Recycle-GAN converts a sequence of frames from one domain to another while maintaining the style of the second domain. A general object retargetting model is trained on only the source domain, and the target domain video is provided at testing time. In contrast, Recycle-GAN contains two GANs frameworks trained on two datasets from the source and target domains. Unlike previous video retargetting models [68, 69, 72, 74], Recycle-GAN is conditioned on videos from one domain, and not on semantic maps or detected poses. Recycle-GAN is an extension of Cycle-GAN[77], which is an unsupervised method used in the image domain to translate unpaired images form one domain to another by applying the cycle consistency loss. Additionally, Recycle-GAN incorporates spatial and temporal constraints within a GANs framework.

## 5 CONCLUSION

Generative models such as GANs provide promising results in multiple domains including images, videos, audios and texts. Video synthesis is still in the early stages compared to other domains such as images. The current state of the art for video GANs suffers from low quality frames or low number of frames or both. One reason could be the higher requirement for computational power as videos are high dimensional data, and so necessitate networks with a large number of parameters. To handle such data, there is a need for a complex architecture that takes into consideration spatial as well as temporal data. For example, DVD-GAN uses one TPU and TGANv2 utilizes 8 GPUs. In addition, videos are usually multimodal and may include audio stream as well, which makes the processing even more complex. Collecting domain-specific videos is also more time-consuming and expensive comparing to other domains such as images, as automatic video retrieval algorithms are not yet very accurate and video data collection involves a lot of manual work to select, clean and pre-process the data. Nevertheless, the trend is upward and every year more studies are being done in this area. The applications of video GANs are broad and include speech animation, video prediction, video retargetting, generating stories from caption and video completion. Although the progress on GANs in areas other than videos is well documented through several review papers, video GANs models have received less attention so far, and if at all included, they were only a section in other review papers despite their broad range. Considering the increasing number of studies on video GANs during the past few years, it is the right time to survey the field, categorise different models according to their applications and compare their differences. This paper is among the initial attempts to review GANs models that produce videos and highlight their main differences.

While 3D CNN GANs as proposed by Vondrick et al. [6] appear to be  an intuitive choice to synthesise videos and represent frames along with time dimension,  3D convolutions may cause overfitting [78]. An alternative to 3D CNN is to utilize RNN with 2D convolution, as in MoCoGAN [40]. Using 2D convolution and 1D convolution to disentangle content from the motion dimension is another way to represent videos [46].

Videos can be generated using GANs either without conditional settings or by introducing a conditional signal such as an audio, image, video, text, label or semantic map. This survey paper initially groups the video GANs frameworks according to their conditional setting: unconditional video generation vs conditional video generation, and discusses the most important models proposed so far in each category and outlines the differences. Moreover, it goes deeper into the conditional frameworks and categorizes the methods according to their condition and presents and compare the different models in each of those categories.  In addition, a list of all datasets used in the reviewed frameworks, their characteristics, evaluation metrics and the loss function applied in each work are presented in supplementary material. The hope is that this paper will serve as a good review of the domain so far and provide the reader with a better understanding of different frameworks and their potential applications.

## A  SUPPLEMENTARY MATERIAL

This supplementary material provides summary tables for the reviewed frameworks in the paper. To better organize the information in different tables and include as much information as possible for each framework, codes linked to Tables S1, S2 and S3 are used. In Table S1, all datasets used in the reviewed papers are listed and coded with prefix "D" and a number (e.g. D1 refers to Clever-sv dataset and so on). It also presents the number of videos in the dataset, duration (if available), resolution and purpose. In Table S2, all metrics used in the reviewed frameworks are summarised and codes with prefix "E" and a number (e.g. E1 refers to human evaluation). Likewise, in Table S3 lists the loss functions used and codes them with prefix L and a number (e.g. L4 refers to GDL loss). In Table S4, the unconditional GANs frameworks reviewed in section 4.1 are listed, and includes the dataset(s), the loss function(s) and the evaluation metric(s) used in each work. In Table S5, speech to video GANs frameworks discussed in section 4.2.1 are summarised. In Table S6, text to video frameworks of section 4.2.2 are listed, and Table S7 is the list of semantic maps to video frameworks in section 4.2.3. In Table S8, all image to video frameworks in section 4.2.4 are listed. Finally, Table S9 is a summary of video to video GANs models of section 4.2.5. All tables contain information such as dataset, type of condition, loss functions and evaluation metrics.

Table S1: Datasets used in video GANs frameworks reviewed in this paper. They are coded with prefix D and number used to link to the other tables. NA means the information for a particular cell is not available

| Abbreviation | Dataset | Purpose | Number of videos | Duration | Scaled Resolution |
|---|---|---|---|---|---|
| D1 | Clevr-sv[39] | layouts of objects | 13000 | NA | NA |
| D2 | UCF-101[79] | human actions | 13320 | NA | 64×64 in [40][44] 192×192 in [47] |
| D3 | Penn action dataset [80] | human actions | 2326 | 32 frames | 64×64 |
| D4 | Kinetics human action [81] | human actions | 500000 | NA | NA |
| D5 | GRID[82] | speakers uttering short phrases | 48614 | NA | NA |
| D6 | TCD TIMIT[83] | speakers uttering short phrases | 9881 | NA | NA |
| D7 | Pororo dataset[84] | cartoon | 16000 | 1 second | NA |
| D8 | President Obama's weekly address videos | Obama uttering speeches | - | NA | NA |
| D9 | BAIR action free[85] | robotic arm pushing a variety of objects | 41216 | NA | 64×64 |
| D10 | FaceForensics[86] | human faces | 854 | NA | 256×256 |
| D11 | Shape motion[40] | two shapes moving | 4000 | 16 frames | 64×64 |
| D12 | Mug facial expression[87] | facial expressions | 3528 | 50-160 frames | 96×96 |
| D13 | Dataset in literature [73] | moving people | - | 12k frames | NA |
| D14 | Tai chi[40] | human actions | 4500 | 32-100 frames | 64×64 |
| D15 | Unlabeled video dataset[6] | golf course, babies in hospital, beaches, and train stations | 2000000 | 32 frames | 64×64 |

| Abbreviation | Dataset | Purpose | Number of videos | Duration | Scaled Resolution |
|---|---|---|---|---|---|
| D16 | Humman3.6M[88] | 3D humans in varies poses | NA | NA | NA |
| D17 | Moving MNIST[89] | moving numbers | 10000 | 16-20 frames | 64×64 |
| D18 | Golf scene [6] | golf scene | 20268 | NA | 64×64 |
| D19 | Sport1M[90] | human actions | NA | NA | 32×32 |
| D20 | THUMOS-15[91] | human actions | NA | NA | NA |
| D21 | NORB: objects[92] | 3D objects | NA | NA | NA |
| D22 | SUNCG: chairs [93] | 3D chair models | NA | NA | NA |
| D23 | Linguistic Data Consortium (LDC) [94] | speakers uttering short phrases | NA | NA | NA |
| D24 | Syn-action dataset [43] | synthetic action | 6000 | NA | NA |
| D25 | Weizmann human action[95] | human actions | 90 | NA | NA |
| D26 | SURREAL[96] | Computer Graphics (CG)human actions | 67582 | 32 frames | 64×64 |
| D27 | LRW [97] | word-level lip reading | 850 | 1 second | 256×256 |
| D28 | CRowd-sourced Emotional Multimodal Actors Dataset (CREMA-D) [98] | speakers uttering short phrases with motion | 7442 | NA | NA |
| D29 | Lip Reading Sentence 3 (LRS3) Dataset[99] | speakers from TED talks | NA | NA | NA |
| D30 | Dataset in literature [42] | human actions | 4000 | NA | NA |
| D31 | KITTI [100] | street scenes | NA | NA | NA |
| D32 | Caltech Pedestrian [101] | street scenes | NA | NA | NA |
| D33 | Dataset(training) in literature [68] | dance | 4 | 8-17 minutes | 1920×1080 1280×720 |
| D34 | Dataset in literature [74] | tennis-walking - fencing | NA | NA | NA |
| D35 | UvA-Nemo[102] | facial dynamics analysis | 1240 | 32 frames | 64×64 |
| D36 | Viper dataset [103] | realistic computer games scenes | 77 | NA | NA |
| D37 | Dataset(training) in literature [69] | dance | 8 | 4-12 minutes | NA |
| D38 | YouTube dancing videos [55] | dance | 1500 | NA | NA |
| D39 | Street-scene videos [56] | street scenes | NA | NA | NA |
| D40 | Cityscapes [104] | street scenes | 5000 | 30 frames | NA |
| D41 | Apolloscape [105] | street scenes | 73 | 100-1000 frames | NA |
| D42 | SBMG[106] | one digit bouncing handwritten | 12000 | 16 frames | 64×64 |
| D43 | TBMG[106] | two digit bouncing handwritten | NA | NA | NA |
| D44 | MSVD [107] | YouTube with manual annotated sentences | 1970 | NA | NA |
| D45 | Shapes-v1 [44] | moving shapes | NA | NA | NA |
| D46 | Shapes-v2 [44] | moving shapes | NA | NA | NA |

| Abbreviation | Dataset | Purpose | Number of videos | Duration | Scaled Resolution |
|---|---|---|---|---|---|
| D47 | Sky scene[60] | time-lapse sky videos | 38207 | 32 frames | 128×128 |
| D48 | Datasets in literature[71] | Varies datasets for public figures giving speeches as Elizabeth II, Putin, and Obama | NA | NA | NA |
| D49 | CelebV [75] | public figures giving speeches | 200000 | 30 minutes | NA |
| D50 | Mixamo | dance | NA | NA | NA |
| D51 | Solo-Dancer[70] | dance | NA | NA | NA |

Table S2: All the evaluation measures used in video GANs frameworks reviewed in this paper. Each evaluation metric is coded with prefix "E" and a number used to link to the other tables.

| Abbreviation | Measures |
|---|---|
| E1 | Human evaluation |
| E2 | Average Content Distance (ACD) |
| E3 | Inception score |
| E4 | Motion control score |
| E5 | Fréchet Inception Distance (FID) |
| E6 | Generative Adversarial Metric (GAM) |
| E7 | Peak Signal-to-Noise Ratio (PSNR) |
| E8 | Structural Similarity (SSIM) index |
| E9 | Cumulative Probability Blur Detection (CPBD) measure |
| E10 | Frequency domain Blurriness Measure (FDBM) |
| E11 | Word Error Rate (WER) |
| E12 | Segmentation accuracy |
| E13 | Pose error |
| E14 | Landmark Distance (LMD) |
| E15 | Classification accuracy |
| E16 | Sharpness measure |
| E17 | Inter-Entropy |
| E18 | Intra-Entropy |
| E19 | Learned Perceptual Image Patch Similarity (LPIPS) |
| E20 | Average VGG cosine similarity |
| E21 | Mean Square Error (MSE) |
| E22 | Maximum Mean Discrepancy (MMD) |
| E23 | Average key point distance |

| Abbreviation | Measures |
|---|---|
| E24 | Missing key point rate |
| E25 | Intersection over Union (IoU) |
| E26 | Mean Pixel accuracy (MP) |
| E27 | Facial action consistency |
| E28 | Mean Absolute Error (MAE) |

Table S3: Loss functions used in video GANs frameworks in the reviewed studies. Each loss function is coded with prefix L and a number used to link to the other tables.

| Abbreviation | Losses |
|---|---|
| L1 | L1 |
| L2 | L2 |
| L3 | Adversarial loss |
| L4 | GDL loss |
| L5 | KL loss |
| L6 | Perceptual loss |
| L7 | Pixel-like loss |
| L8 | VAE loss |
| L9 | Negative log-likelihood function |
| L10 | Cosine similarity loss |
| L11 | Feature-matching loss presented in pix2pixHD |
| L12 | Mask loss |
| L13 | Recycle-loss |
| L14 | Recurrent loss |
| L15 | Semantic layout and pose feature losses |
| L16 | Margin ranking losses |
| L17 | Flow loss |
| L18 | Forward-backward consistency loss |
| L19 | Reconstruction loss |
| L20 | Cycle loss |
| L21 | Triplet Loss |
| L22 | Ranking Loss |

## A.1 Unconditional GANs

Table S4: The reviewed unconditional video GANs frameworks in section 4.1. The second column provides the type of condition. The third, fourth and fifth columns give information on loss functions, training datasets and evaluation measures respectively. More info

| Publication | Conditional information | Loss | Dataset | Measures |
|---|---|---|---|---|
| VGAN[6] | no condition | L3 | D15 | E1, E15 |
| FTGAN[45] | no condition | L3 | D3, D26 | E1, E15 |
| MoCoGAN[40] | no condition | L3 | D2, D11, D12, D14, D25 | E1, E2, E3, E4 |
| TGAN [46] | no condition | L3 (WGAN) | D2, D17, D18 | E1, E3, E6 |
| TGANv2 [47] | no condition | L3 | D2, D10 | E3, E5 |
| DVD-GAN [48] | no condition | L3 | D2, D4 | E3, E5 |

## A.2 Conditional GANs

Table S5: The reviewed speech to video GANs frameworks in section 4.2.1. The second column provides the type of condition. The third, fourth and fifth columns give information on loss functions, training datasets and evaluation measures respectively. More inform

| Publication | Conditional information | Loss | Dataset | Measure |
|---|---|---|---|---|
| Vougioukas et al.[50] | audio, initial image | L1, L3 | D5, D6 | E1, E2, E7 -11 |
| DAVS [51] | (audio or video), initial image | L1, L3 | D27 | E1, E7, E8 |
| Mittal et al.[52] | audio-based content representations, initial image | L1, L2, L3, L5, L6, L9, L16 | D5, D28, D29 | E7, E8, E14 |
| Chen et al.[49] | audio, initial image | L3, L6, L7, L10 | D5, D23, D27 | E7-9 |
| Jalalifar et al.[53] | audio, set of landmarks | L3 | D8 | - |

Table S6: The reviewed text to video GANs frameworks in section 4.2.2. The second column provides the type of condition. The third, fourth and fifth columns give information on loss functions, training datasets and evaluation measures respectively. More information on the last three columns can be found in Tables S1-S3.

| Publication | Conditional information | Loss | Dataset | Measure |
|---|---|---|---|---|
| TGANs-C[54] | one sentence | L2, L3 | D42, D43, D44 | E1, E6 |
| Li et al.[42] | one sentence | L1, L3, L8 | D30 | E15 |
| TFGAN [44] | one sentence | L3 | D4, D45, D46 | E5, E15 |
| StoryGAN [39] | Multiple sentences | L3, L5 | D1, D7 | E1, E8, E15 |

Table S7: The reviewed semantic map to video GANs frameworks in section 4.2.3. The second column provides the type of condition. The third, fourth and fifth columns give information on loss functions, training datasets and evaluation measures respectively. More information on the last three columns can be found in Tables S1-S3.

| Publication | Conditional information | Loss | Dataset | Measure |
|---|---|---|---|---|
| Vid2vid [55] | semantic video | L3, L17 | D10, D38, D40, D41 | E1, E5 |
| Few-shot-vid2vid [56] | Semantic video, initial image | L3, L17 | D10, D38, D39 | E1, E5, E12, E13 |
| Pan et al.[57] | semantic video, initial image | L1, L5, L6, L18, L19 | D2, D4, D31, D40 | E1, E5 |

Table S8: The reviewed image to video GANs frameworks in section 4.2.4. The second column provides the type of condition. The third, fourth and fifth columns give information on loss functions, training datasets and evaluation measures respectively. More information on the last three columns can be found in Tables S1-S3.

| Publication | Conditional information | Loss | Dataset | Measures |
|---|---|---|---|---|
| Mathieu et al.[58] | input frames | L2, L3, L4 | D2, D19 | E7, E8, E16 |
| Lee et al.[59] | input frame | L3, L8 | D4, D9 | E1, E7, E8, E20 |
| MCnet [63] | input frames | L2, L3, L4 | D2, D4, D25 | E7, E8 |
| Walker et al.[64] | input frames | L3, L8 | D2 | E3, E22 |
| TwoStreamVAN[43] | input frame, category, motion map | L3, L8 | D12, D24, D25 | E3, E15, E17, E18 |
| Liang et al.[65] | input frames | L3, L8 | D2, D20, D31, D32 | E7, E8, E21 |
| DRNET [66] | input frames | L2, L3 | D4, D17, D21, D22 | E15, E3, |
| Hu et al. [67] | input frame, stroke | L2, L3, L6 | D4, D16, D17 | E5, E19 |
| MD-GAN [60] | Input frame | L1, L3, L22 | D47 | E7, E8, E21 |
| Cai et al.[61] | Input image, label | L1, L2, L3 | D2, D16 | E3 |

Table S9: The reviewed video to video GANs frameworks in section 4.2.5. The second column provides the type of condition. The third, fourth and fifth columns give information about loss functions, training datasets and evaluation measures respectively. More information on the last three columns can be found in Tables S1-S3.

| Publication | Conditional information | Loss | Dataset | Measure |
|---|---|---|---|---|
| Zhou et al.[69] | video, image | L3, L6, L11, L15 | D37 | E1, E7, E8, E21 |
| Chan et al. [68] | video | L3, L6, L11 | D33 | E1, E8, E19 |
| Vid2Game[74] | video, control signal | L3, L6, L11, l12 | D34 | E8, E19 |
| Monkey-net [72] | video, image | L3, L11 | D9, D14, D35, | E1, E5, E21, E23, E24 |
| Recycle-GAN [76] | video | L3, L13, L14 | D36 | E1, E15, E25, E26 |
| Liu et al. [73] | video, image | L1, L3 | D13 | E1, E8 |
| Deep Video Portraits [71] | video | L1, L3 | D48 | E1, E28 |
| ReenactGAN [75] | video | L1, L2, L3, L20 | D49 | E1, E27 |
| TransMoMo[70] | video | L1, L3, L19, L21 | D50, D51 | E1, E3, E21, E28 |